\documentclass[10pt,twocolumn,letterpaper,xcolor={dvipsnames}]{article}

\usepackage{iccv}
\usepackage{times}
\usepackage{epsfig}
\usepackage{graphicx}
\usepackage{amsmath}
\usepackage{amssymb}
\usepackage{enumitem}

\usepackage{amssymb}% http://ctan.org/pkg/amssymb
\usepackage{pifont}% http://ctan.org/pkg/pifont
\newcommand{\cmark}{\ding{51}}%
\newcommand{\xmark}{\ding{55}}%

\usepackage{multirow}
\usepackage{makecell}
\usepackage{algpseudocode}
\usepackage{mathtools}
\usepackage{centernot}

\usepackage[textwidth=1.8cm]{todonotes}
\newlength{\arrow}
\usepackage{array}
\newcommand{\PreserveBackslash}[1]{\let\temp=\\#1\let\\=\temp}
\newcolumntype{C}[1]{>{\PreserveBackslash\centering}p{#1}}
\newcolumntype{R}[1]{>{\PreserveBackslash\raggedleft}p{#1}}
\newcolumntype{L}[1]{>{\PreserveBackslash\raggedright}p{#1}}

\usepackage[ruled,vlined]{algorithm2e}
\definecolor{commentcolor}{RGB}{110,154,155}
\definecolor{darkergreen}{RGB}{0,153,0}% define comment color
\newcommand{\PyComment}[1]{\ttfamily\textcolor{commentcolor}{\# #1}}  % add a "#" before the input text "#1"
\newcommand{\PyCode}[1]{\ttfamily\textcolor{black}{#1}} % \ttfamily is the code font

\newcommand{\ours}{{$\mathtt{LatentDR}$}} 

\usepackage[pagebackref=true,breaklinks=true,letterpaper=true,colorlinks,bookmarks=false]{hyperref}

\iccvfinalcopy % *** Uncomment this line for the final submission

\def\iccvPaperID{10334} % *** Enter the ICCV Paper ID here

\ificcvfinal\fi

\begin{document}

%%%%%%%%% TITLE
\title{
% Sample-aware latent degradation and restoration for domain generalization
\Large LatentDR: Improving Model Generalization Through\\ Sample-Aware Latent Degradation and Restoration
% LatentDR: Sample-Aware Latent Degradation and Restoration \\ for Out-of-Distribution (OOD) Generalization
}

% \author{Ran Liu\\
% Georgia Tech\\
% %Institution1 address\\
% {\tt\small rliu361@gatech.edu}
% % For a paper whose authors are all at the same institution,
% % omit the following lines up until the closing ``}''.
% % Additional authors and addresses can be added with ``\and'',
% % just like the second author.
% % To save space, use either the email address or home page, not both
% \and
% Sahil Khose\\
% Georgia Tech\\
% %First line of institution2 address\\
% {\tt\small sahil.khose@gatech.edu}
% \and
% Jingyun Xiao\\
% Georgia Tech\\
% %First line of institution2 address\\
% {\tt\small jxiao76@gatech.edu}
% \and
% Lakshmi Sathidevi
% \\
% Georgia Tech\\
% %First line of institution2 address\\
% {\tt\small lsathidevi3@gatech.edu}
% \and
% Keerthan Ramnath\\
% Georgia Tech\\
% {\tt\small kramnath6@gatech.edu}
% \and
% Zsolt Kira\\
% Georgia Tech\\
% %First line of institution2 address\\
% {\tt\small  zkira@gatech.edu}
% \and
% Eva L. Dyer\\
% Georgia Tech\\
% {\tt\small evadyer@gatech.edu}
% }

\author{\large Ran Liu, 
 Sahil Khose,
Jingyun Xiao,
Lakshmi Sathidevi,
Keerthan Ramnath,
Zsolt Kira,
Eva L. Dyer\\
\large Georgia Institute of Technology, Atlanta, GA, 30332
}

\maketitle
% Remove page # from the first page of camera-ready.
\ificcvfinal\thispagestyle{empty}\fi

%%%%%%%%% ABSTRACT
\begin{abstract}
Despite significant advances in deep learning, models often struggle to generalize well to new, unseen domains, especially when training data is limited. To address this challenge, we propose a novel approach for distribution-aware latent augmentation that leverages the relationships across samples to guide the augmentation procedure. Our approach first degrades the samples stochastically in the latent space, mapping them to augmented labels, and then restores the samples from their corrupted versions during training. This process confuses the classifier in the degradation step and restores the overall class distribution of the original samples, promoting diverse intra-class/cross-domain variability. We extensively evaluate our approach on a diverse set of datasets and tasks, including domain generalization benchmarks and medical imaging datasets with strong domain shift, where we show our approach achieves significant improvements over existing methods for latent space augmentation. We further show that our method can be flexibly adapted to long-tail recognition tasks, demonstrating its versatility in building more generalizable models. Code is available at \href{https://github.com/nerdslab/LatentDR}{https://github.com/nerdslab/LatentDR}.
% Our results demonstrate that our approach outperforms existing state-of-the-art methods in terms of out-of-domain generalization and long-tail recognition.
\end{abstract}% \zkn{Great abstract!}

%%%%%%%%% INTRO

\section{Introduction}
In machine learning, it is often challenging to train a model that generalizes well when tested on domains that it was not exposed to during training \cite{ben2010theory}. This is especially true when the available training data is limited, as is often the case in real-world applications. Various methods have been proposed to address this challenge, including feature alignment approaches \cite{sun2016deep,ganin2016domain}, meta-learning \cite{sankaranarayanan2023meta,balaji2018metareg}, and data augmentations \cite{volpi2018generalizing,zhou2020deep}. However, achieving robust out-of-domain (OOD) generalization remains a challenging problem \cite{gulrajani2020search}, and further research is needed to develop more effective methods.

\begin{figure}[t!]
\centering
  \includegraphics[width=0.45\textwidth]{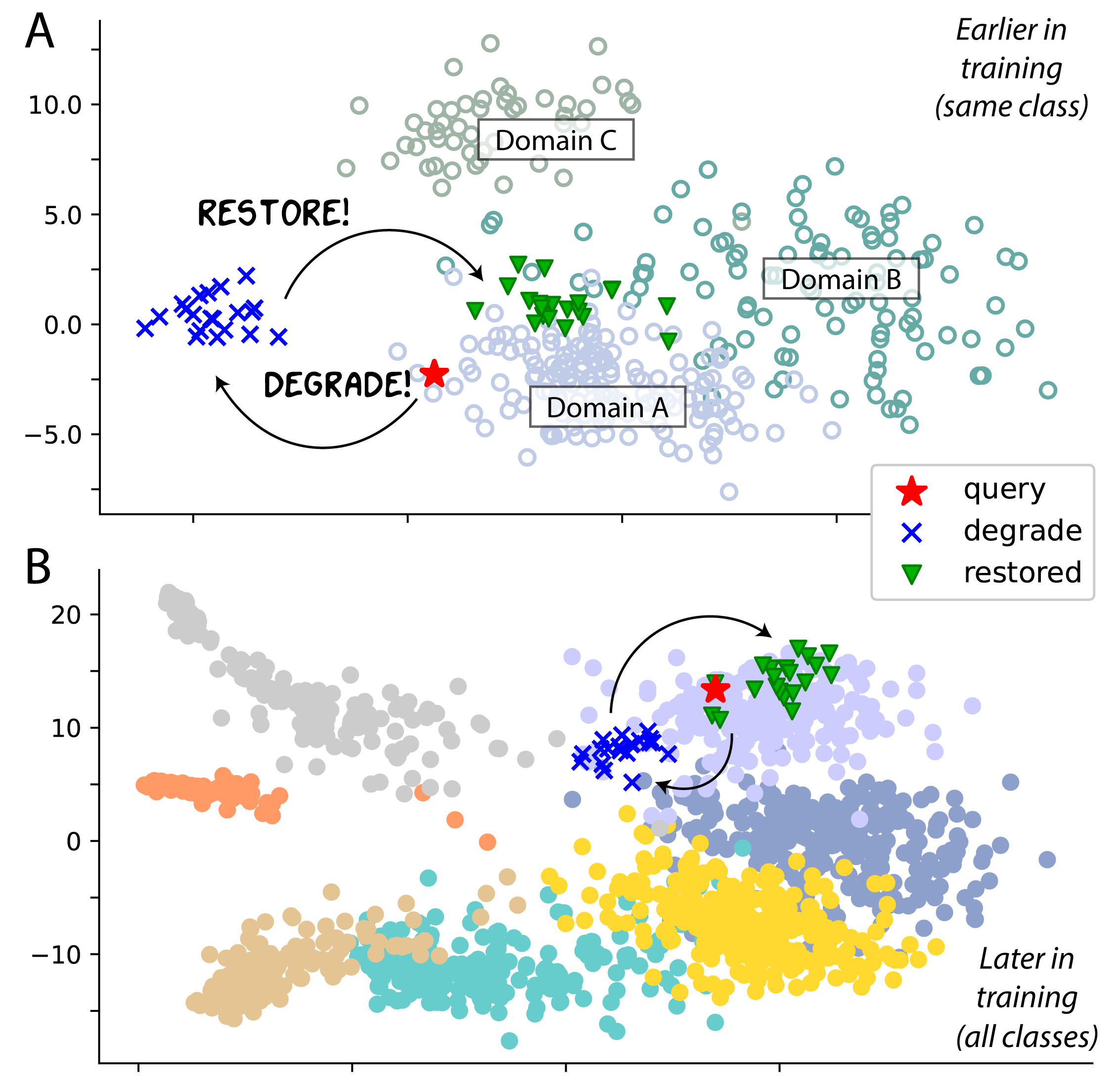}
  \vspace{-2mm}
  \caption{\small {\em Visualization of latent degradation and restoration steps during training.} In \textbf{(A)}, when it is earlier in training, latents from different domains are separated for the same class, which hurts model generalization. To tackle this issue, for each {latent query} ({\color{red}{$\bigstar$}}), \ours~first {degrades} ({\color{blue}{\bf$\mathsf{x}$}}) them to a point that is far away from the existing samples, and {restores} ({\color{darkergreen}{$\blacktriangledown$}}) them back to the existing distribution to diversify existing domains and improve model generalization. In \textbf{(B)}, 
  we visualize the latents for all classes at a later point in training and color data by classes. More details and visualizations are in Appendix~\ref{appendix:latents}.\label{fig:fig1}\vspace{-5mm}}
\end{figure}

Among existing strategies, one emerging technique for addressing the problem of OOD~generalization is {\em latent augmentation}.
Similar to data augmentation methods, the overall objective of latent augmentation is to increase the diversity of the source data so that the model is encouraged to learn domain-agnostic representations \cite{verma2019manifold}. However, different from data augmentation methods which directly manipulate the input data, 
latent augmentation modifies the hidden data representations or feature space of a model, which avoids the usage of generative networks \cite{zhou2020deep}, additional classifiers \cite{shankar2018generalizing}, or adversarial training \cite{volpi2018generalizing}. %[Add some positive part about latent augs]

Many existing latent augmentation methods typically assume that linear combinations \cite{verma2019manifold}, the mixture of style statistics \cite{hong2021stylemix,somavarapu2020frustratingly}, or additional randomization \cite{li2021simple} of hidden representations can provide the desired diversity for generalization. However,
when dealing with highly diverse domains, using simple assumptions (like linear mixing) is often insufficient \cite{cha2021swad}, and generating robust latent augmentations remains an outstanding challenge. Here we ask if we can go beyond pairwise mixing and use sample-to-sample relationships across points in the latent space to build sample-aware latent augmentations that can allow us to transport across domains.

To build sample-aware latent augmentations, we take inspiration from recent work where batch-level relationships are learned through an attention mechanism to ``reconstruct'' the original latent samples \cite{hou2022batchformer}. However, rather than trying to reconstruct the original sample, we consider a bi-directional objective where the goal is to both create a \textit{degraded} sample that confuses a downstream classifier and also create a \textit{restored} sample that recovers the class information after this transformation. In both steps, we use relationships across many samples within mini-batches to help build augmentations. %and solve the problem of restoration. 
The idea is shown in Figure~\ref{fig:fig1}(A): The degradation process aims to map a query latent to a point where its class cannot be easily distinguished by a downstream classifier; 
the 
restoration process uses a 
cross-attention mechanism \cite{vaswani2017attention,jaegle2021perceiver} to 
(only) recover the class information from the degraded latents by conditioning on the original latents. 
By combining degradation and restoration, \ours~encourages the model to capture the relationship across samples from different classes and domains, and thus guides the model to generalize better to unseen sources.

To test our approach in diverse settings, we conduct experiments on: (i) five standardized domain generalization benchmarks \cite{gulrajani2020search,cha2021swad},
(ii) five medical imaging classification datasets that suffer from 
domain shift \cite{koh2021wilds,yang2023medmnist}, and (iii) 
a long-tail (LT) recognition task where 
our latent augmentation technique can improve generalization in the presence of strong data imbalance.
In all of these cases, we provide impressive boosts over other augmentation-based approaches on all of the datasets tested, and competitive performance with state-of-the-art methods that use more complex losses and domain information. To better understand the effectiveness of our approach on learning, we use two measures of representation quality \cite{wang2020understanding} and nearest neighbor visualization to measure the impact of our approach.

In summary, the contributions of our work include:
\begin{itemize}[itemsep=0.2pt,topsep=0.2pt] %noitemsep can complete remove space between items
\item We propose a novel approach for latent augmentation that aims to build connections across samples from different domains by both \textit{degrading} and \textit{restoring} the samples from the latent distribution.

\item We conduct empirical studies to demonstrate that our approach improves class-level alignment of features, as well as the uniformity or spread of the data distribution. 
These two properties are shown to be strongly correlated with downstream task performance \cite{wang2020understanding}.

\item We extensively evaluate the robustness of our method on domain generalization benchmarks and medical imaging datasets with strong domain shifts, where we provide various ablations to better understand different components of the method. Finally, we further demonstrate the versatility of our augmentation method by applying it to a long-tail recognition task.

\end{itemize}

%\zkn{Much better intro! Just needs a bit more precise language tying some of the claims (e.g. sample-to-sample relationships) to elements of the method. Maybe a good framing is: 1) Inspired by diffusion, we propose a degradation/restoration process, 2) Addressing limitations of simple sample-to-sample mixing in prior works, we leverage batch-level sample-to-sample mixing through self-attention to perform this degradation/restoration, 3) Crucially, the restoration process is conditioned on the non-corrupted latents, using sample-to-sample relationships including to non-corrupted latents. This story is kind of there, but without the flow and not as explicitly/directly stated in many places. }

%\input{texts/intro_feb28}
%\input{texts/intro_march4}

%%%%% BACKGROUND

\section{Background and Related Work}

\subsection{Domain generalization (DG)}

Various methods have been proposed to address the challenge of domain generalization, including applying explicit domain alignment \cite{ganin2016domain,sun2016deep}, 
% wang2020domainmix
learning domain-agnostic representations \cite{nam2021reducing,huang2020self,zhao2020domain}, meta-learning \cite{balaji2018metareg,bui2021exploiting,sankaranarayanan2023meta}, enforcing domain invariance by adjusting direction of the gradients
%optimzing the direction of the gradient-related approaches 
\cite{shankar2018generalizing,shi2021gradient,rame2022fishr}, and data or latent augmentation methods \cite{hong2021stylemix,somavarapu2020frustratingly,li2021simple}. An orthogonal line of research aims to develop better optimization approaches to alleviate the overfitting issue \cite{cha2021swad,arjovsky2019invariant,krueger2021out}, which often happens in DG settings.
More recently, a line of work focuses on taming large-scale pre-trained models to improve domain generalization performance through oracle methods \cite{cha2022domain,li2022domain} or prompts \cite{niu2022domain,zheng2022prompt}. %\zkn{Might be good to emphasize the part about ``more complex losses and domain information'' mentioned in intro}. 
 Despite the numerous proposed approaches, domain generalization remains a challenging problem, with only a few methods consistently outperforming Empirical Risk Minimization (ERM) with properly designed data augmentation and training protocols \cite{gulrajani2020search}.

\subsection{Data and latent augmentation for DG}
Augmentation approaches have been extensively used to improve the generalization performance of models \cite{balestriero2022effects,lin2022good}. Commonly used image-space data augmentation methods for DG include general-purpose operations \cite{zhang2017mixup,devries2017improved,yun2019cutmix,kim2020puzzle},
and DG-specific methods \cite{zhou2020deep,shankar2018generalizing,volpi2018generalizing,zhou2020learning,sankaranarayanan2023meta}.
Due to the additional complexity and high computation demand of operating directly on data space,
recent works focus on exploring the use of latent space augmentation methods \cite{verma2019manifold}, which aim to manipulate data in the feature space. 
For instance, Zhou et al. \cite{zhou2021domain} and Somavarapu et al. \cite{somavarapu2020frustratingly} proposed to perform linear mixing of style statistics with AdaIN module \cite{huang2017arbitrary}, while Li et al. \cite{li2021simple} and Wang et al. \cite{wang2022feature} introduced randomization modules to augment the latent space. However, many of these models operate on within-domain datasets and few studies have demonstrated their scalability in large-scale DG experiments \cite{gulrajani2020search}.

\vspace{-4mm}
% \paragraph{Sample-aware approaches for latent augmentation.}~
\paragraph{Sample-aware approaches.}~
Augmentation methods typically rely on sample-to-sample relationships to improve model generalization, such as Mixup which interpolates between pairs of samples. Recently, non-convex methods have been proposed to learn more sophisticated sample-to-sample relationships. For example, Batch-Graph \cite{wang2022rethinking} models relationships using a graph, while BatchFormer \cite{hou2022batchformer, hou2022batchformerv2} uses self-attention to encourage gradient flow between samples in a batch. Building on these works, \ours~also models sample-to-sample relationships using attention, but with the goal of degrading and restoring the latents using information from other samples in the batch.
%\zkn{Great description! This description that you leverage samples across the batch  should be in intro (batches are never even mentioned)}. 
By leveraging the relationships across samples, our method aims to confuse the classifier during degradation and restore the original class distribution during restoration, ultimately improving the model's ability to generalize to new domains.

\subsection{Transformers}

While transformers were initially proposed in the context of natural language processing applications for learning sequences \cite{vaswani2017attention}, the use of transformers and related components is becoming commonplace in vision applications \cite{dosovitskiy2020image,liu2022convnet}, augmentation \cite{stegmuller2023scorenet}, and conditioning \cite{rombach2022high}.

\begin{figure*}[t]
\centering
  \includegraphics[width=0.95\textwidth]{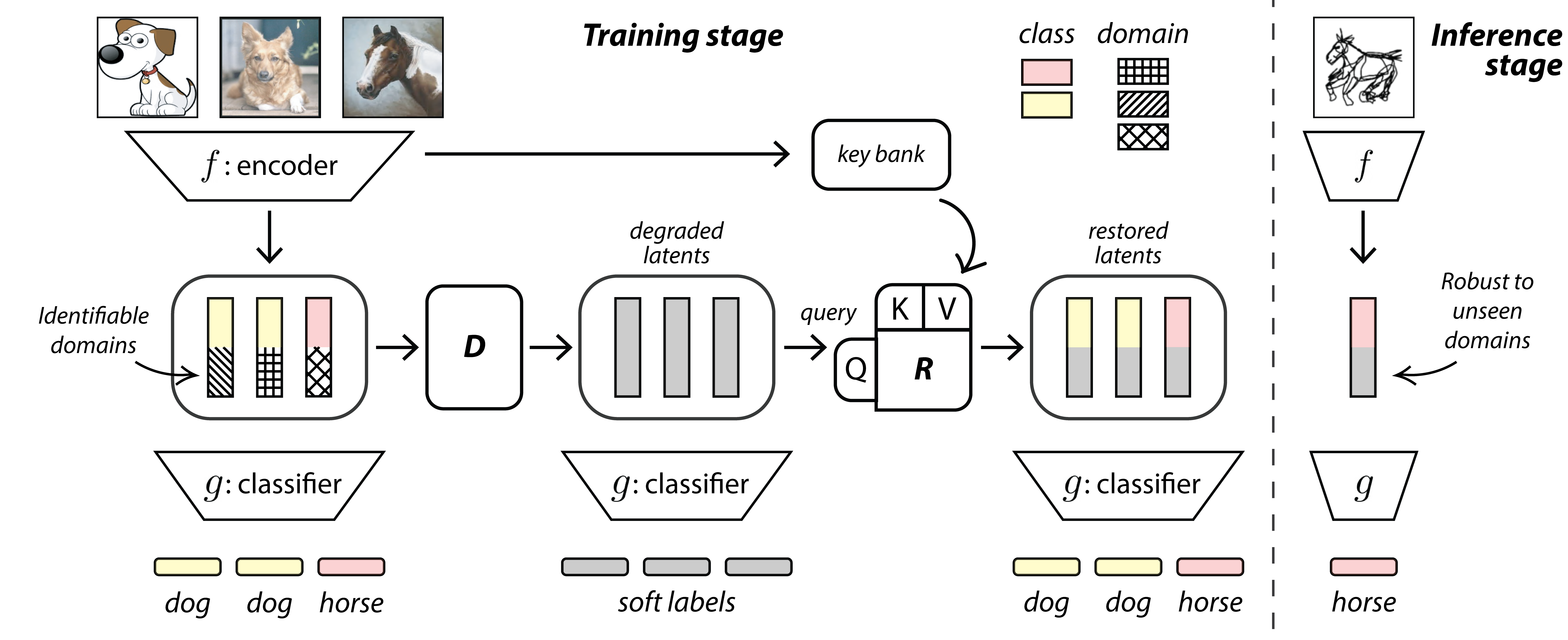}
  \caption{\textit{Detailed architecture of LatentDR.} 
  \textbf{(A)} Without latent augmentation, latents produced by encoders would contain identifiable information about domains, which hurts the generalization ability of the model. To address this issue, \ours~first uses a degradation operator to produce degraded latents that are mapped to constructed soft labels. Subsequently, we fed degraded latents as queries and original latents as keys to a cross-attention transformer, which restores latents to assorted domains. \textbf{(B)} Applying augmentations during training improves the model's ability to generalize, resulting in latent features that are robust to unseen domains, even when the degradation and restoration operators are removed during inference.
  % During inference, the latent augmentations make the produced latent robust to unseen domains, even when the degradation and restoration operators are removed.
  \label{fig:arch}\vspace{-2mm}
  }
\end{figure*}

\vspace{-3mm}
\paragraph{Transformer with self-attention.}
A typical transformer layer processes a sequence of inputs $X \in \mathbb{R}^{N \times d}$ through the multi-head self-attention (MSA) mechanism to learn the relationship among
$N$ tokens of $d$-dimensions. For each token, we derive its queries $Q$, keys $K$, and values $V$ using linear projections, and perform the following operations:
\begin{gather}
Q = XW_Q, K = X W_K , V = X W_V \\
\operatorname{Attention}(Q,K,V) = \operatorname{softmax}\left( {Q K^{T}/\sqrt{d_k}}\right) V,
\end{gather}
where $\operatorname{softmax}(\cdot)$ gives row-wise softmax normalization.

The transformer architecture divides self-attention into multiple different heads, and combines the attention mechanism (MSA), MLP blocks (FF), and layerwise normalization (LN). A transformer layer consists of below operations:
\begin{equation}
\label{eq:MSA}
\begin{aligned}
    & Z_{\ell+1}^{\prime} = \operatorname{LN}(Z_{\ell} + \operatorname{MSA}(Z_{\ell})) \\
    & Z_{\ell+1} = \operatorname{LN}(Z_{\ell+1}^{\prime} + \operatorname{FF}(Z_{\ell+1}^{\prime})), ~0 \leq \ell \leq L-1
\end{aligned}
\end{equation}
where $Z_{\ell}$ is the output of the $l$-th layer of the transformer.

\vspace{-3mm}
% \paragraph{Conditioning with cross attention:}
\paragraph{Study correlation across tokens with spatial operations.}
%\zkn{``Study correlation with spatial operation'' is awkward, maybe change to just ''Spatial correlation across tokens''. }\zkn{Also, why is this relevant? Seems like a minor detail unless it's somehow relevant to the method}
Recent works have shown that self-attention can be replaced with a spatial MLP \cite{tolstikhin2021mlp}, or even a Pooling operation \cite{yu2022metaformer}, to study the spatial correlations across tokens. In the case of vision transformers \cite{dosovitskiy2020image}, where each token represents a patch in the image, the pooling operator simply takes an average or max pooling over subsets of embeddings for spatially-correlated (nearby) patches.

\vspace{-3mm}
\paragraph{Conditioning with cross-attention.}
Whereas self-attention considers the pairing of queries and keys across tokens in the same sequence, cross-attention (CA) can be used to compute attention between two different input sources or modalities \cite{vaswani2017attention,lee2019set,jaegle2021perceiver,rombach2022high}:
Given $X_1 \in \mathbb{R}^{N \times d_1}$ and $X_2 \in \mathbb{R}^{N \times d_2}$, the queries $Q$, keys $K$, and values $V$ are computed from different input sources:
\begin{gather}
Q = X_1W_Q, K = X_2 W_K , V = X_2 W_V
\end{gather}
The permutation-invariant property of the cross attention mechanism allows the generated token to be dependent on the queries ($X_1$) while keeping the information from the conditions ($X_2$).
%From the perspective of content/style separation, the output would have the `content' of queries ($X_1$) while keeping the `style' of the conditions ($X_2$).\zkn{This last sentence is hard to understand without more description/context (you don't describe what content/style is, how it relates to DG, etc. Maybe add context, mention later in method, or remove (I would vote for latter unless prior works have use CA in this way, given that this is a background section.}

%%%%% METHOD
\section{Method}

% \subsection{Background}

\subsection{Overview of our method}
To capture relationships across samples and build robustness into the model,
our latent augmentation operation is learned through a {\em degradation} (D) step and a {\em restoration} (R) step, hence the name \ours.
% \footnote{The name LatentDR can also refer to a Latent Doctor that aims to heal the latent space through restoration.\zk{Interesting though not sure you want to propose two different namings :) }}. 
We visualize both steps in Figure~\ref{fig:fig1} and provide an overview of the architecture in Figure~\ref{fig:arch}.
The method does not require layerwise manipulations, does not rely on domain information, and can be implemented easily within the mini-batch training process and removed during inference.

Consider $(x, y)$ as an original datapoint and label pair from our data distribution $(\mathcal{X}, \mathcal{Y})$. Let $z = f(x)$ denote the embedding of $x$ after the encoder $f: \mathcal{X} \rightarrow \mathbb{R}^d$, and $g: \mathbb{R}^d \rightarrow \mathcal{Y}$ as the classifier that produces a final prediction.
The overall objective of the method is to learn a latent degradation operator $D_{\mu}(; \xi)$ that produces stochastic ($\xi$) embeddings that have the potential to `confuse' a downstream classifier; and a latent restoration operator $R_{\theta}$ that would `denoise' the prediction of the labels.

This high-level objective can be specified as:
\begin{equation}
\begin{aligned}
\text{\textit{degrade:} } \quad & \min _{f, g, \mu} \ell(g(D_{\mu}(z; \xi)), \tilde{y}) \\
\text{\textit{restore:} } \quad & \min _{f, g, \theta, \mu} \ell(g(R_\theta(D_{\mu}(z; \xi))), y)
\end{aligned}
\end{equation}
where $\ell$ is a  classification loss, $\tilde{y}$ is a distribution-aware label that is constructed to guide mixing in latent space (detailed in Section~\ref{subsec:latentdr}), $\theta$ and $\mu$ denotes the weights of restoration and degradation operators, and $\xi$ denotes the stochasticity or randomness in the degradation process.
% that is constructed with the intent of generating an augmentation that will confuse . We will provide further details on how to construct these mappings in the next subsection.
%\zkn{Need to describe what this means (I know it's detailed later so succinct summary/description)}.\zkn{What is $\mu$?}
%Intuitively, this 
% can be viewed as 
%is analogous to
%an adversarial\zkn{Do we want to use the term adversarial? I think there are explicit adversarial methods and your method is not quite the same (unless I'm mistaken). } process: 
Hence, while the restoration operator learns to predict the original distribution of labels, the degradation operator aims to introduce non-trivial perturbation to the latent variables to alter the classifier's prediction.
We provide more motivations in Appendix~\ref{appendix:alg-details}.
Note that, different from other methods \cite{li2018domain1}, our recovered latent $z_{r} = R_{\theta}(D_{\mu}(z; \xi))$ does not necessarily need to be close to its original position $z$ as long as the classifier's prediction is correct, which encourages the latents to go across domains for domain generalization.

\subsection{LatentDR}
\label{subsec:latentdr}

\subsubsection{Latent degradation with soft-labels}
\label{subsubsec:d}

In order to encourage the datapoints from different sources to mix with each other,
%and provide additional sample information during training, 
% we augment the latents and generate their corresponding pseudo-labels.
we rely on \textit{sample-to-sample relationships}
% \zkn{Mention batch aspect to make it clear how. This is not clear at all at this point. } 
between points to construct the latent degradation operator $D_{\mu}(;\xi)$ and the corresponding soft-labels $\tilde{y}$.

Let $(\mathcal{S}_X, \mathcal{S}_Y) = \{ (x_i, y_i) \}_{i=1}^B$
% \zkn{Define $B$ as batch size}
denote a set of samples from $(\mathcal{X}, \mathcal{Y})$. % and their corresponding labels.
Let $Z_{set}\in \mathbb{R}^{B \times d}$ 
denote a matrix containing the embeddings of the samples in $\mathcal{S}_X$ and let $z = f(x)$ be a {\em query} latent vector. In practice, one convenient way is to draw a query from the batch and use the remaining samples as $Z_{set}$. % However, we can also consider other variants with our general formulation. 
%\zkn{Where does this come from? Is it an element of $Z_set$?}.
The goal is to generate % $(z_d, y_d) = 
$(\operatorname{Mix}_{\text{data}}(z, Z_{set}), \operatorname{Mix}_{\text{label}}(y, Y_{set}))$
where $\operatorname{Mix}$ is a general (and potentially nonlinear) mixing operator that combines samples and labels within the set to build an augmentation of the query $z$ and its label $y$.
We can consider latent Mixup and other pairwise mixture schemes as a special instance where $Z_{set}$ consists of only one sample and the combination operator is (typically) 
linear for both data and label.

To create non-linear and non-trivial sample-aware augmentations,
we rely on transformers similar as in \cite{hou2022batchformer,hou2022batchformerv2}.
% study two different choices for sample-level augmentation, 
Specifically, in this work, we tested two variants of transformers, where the first uses self-attention (SA) and the other a pooling mechanism (Pool). Thus, we can generate the degraded augmentation $z_d$ of a query $z$ as follows: 
\begin{equation}
\begin{aligned}
(SA) &\quad z_d^{\prime} = z + \operatorname{AttN}(Z_{set}), \\
(Pool) &\quad z_d^{\prime} = z + \operatorname{Pool}( \Omega (Z_{set})), \\
&\quad z_d = z_d^{\prime} + \operatorname{MLP}(z_d^{\prime})
\end{aligned}
\end{equation}
where $\Omega (\cdot)$ is a spatial selection operator, and $\operatorname{Pool}(\cdot)$ takes an average 
pooling over the selected subset. In practice, we also apply normalization (see Appendix~\ref{appendix:alg-details}).
The stochasticity (parameterized by $\xi$ in our degradation operator) is introduced through a high dropout rate of $50\%$ in both operations, and additionally by taking a different subset of samples in the Pooling variant.

To encourage the degraded latent to be apart from the query,
we define the mixing operator for our labels as:
\begin{equation}
\widetilde{y} = \sum_{y_i \in \mathcal{S}_Y } w_i y_{i}.
\end{equation}
where we take $w_i=1/B$ for simplicity.
Note that $Z_{set}$ fits perfectly into mini-batch training, such that it can just be a batch of data that produces degraded latent samples $Z_{d}$.

\subsubsection{Latent restoration with cross-attention}~
To encourage the model to preserve critical semantic information during the degradation process, we perform \textit{latent restoration} to recover
the latents back to their original classes. Our model leverages the cross-attention mechanism that computes queries $Q$ from the corrupted latents $Z_d$ and the keys $K$ and values $V$ from the original latents $Z_{set}$:
%\begin{gather*}
\begin{equation}
Q = Z_d W_Q, K = Z_{set} W_K , V = Z_{set} W_V,
\end{equation}
where we use them to produce the restored latent $z_r$:
\begin{equation}
\begin{aligned}
& z_{r}^{\prime} = z_d + \operatorname{softmax}(QK^T) V \\
& z_{r} = z_{r}^{\prime} + \operatorname{MLP}(z_{r}^{\prime})
\end{aligned}
\end{equation}
%\end{gather*}
As the attention mechanism is permutation-invariant to the ordering of the keys, the output is decided based on the ordering of the queries, which enforces the corrupted latents to contain the information that is sufficient to recover the class information of the original latents.
Thus, intuitively, 
the restoration operator generates a new latent $z_r$ by having $z_d$ to select similar samples from the original latent set $Z_{set}$, which provides assistance in the restoration process.

\begin{table*}
\begin{center}
\begin{tabular}{c|C{0.35cm}C{0.35cm}| l C{1.6cm} C{1.6cm} C{1.6cm} C{1.6cm} C{1.6cm} c}
 & (1) & (2) & Model & PACS & VLCS & OfficeHome & TerraInc & DomainNet & Avg. \\
\hline\hline
\parbox[t]{2mm}{\multirow{7}{*}{\rotatebox[origin=c]{90}{Algorithms}}}
% \multirow{8}{*}{Algorithms} 
 & % \cmark & \xmark
 \cmark & \cmark
 & MMD$^{\dagger}$ \cite{li2018domain1} & 84.7 & 77.5 & 66.4 & 42.2 & 23.4 & 58.8  \\
& % \cmark & \xmark 
\xmark & \cmark
& DANN$^{\dagger}$ \cite{li2018domain2} & 83.7 & 78.6 & 65.9 & 46.7 & 38.3 & 62.6  \\
%&&& RSC$^{\dagger}$ & 85.2 & 77.1 & 65.5 & 46.6 & 38.9 & 62.7 \\
&% \cmark & \cmark 
\cmark & \cmark
& SagNet$^{\dagger}$ \cite{nam2021reducing} & \textbf{86.3}  & 77.8 & 68.1 & \underline{48.6} & 40.3 & 64.2  \\
&% \cmark & \cmark 
\cmark & \cmark
& CORAL$^{\dagger}$ \cite{sun2016deep} & \underline{86.2} & \underline{78.8} & \underline{68.7} & 47.7 & 41.5 & \underline{64.5}  \\
& %\cmark & ? 
\cmark & \cmark
& MLDG$^{\dagger}$ \cite{li2018learning}& 84.9 & 77.2 & 66.8 & 47.7 & 41.2 & 63.6 \\
& %\cmark & \xmark 
\xmark & \cmark
& Fishr$^{\dagger}$ \cite{rame2022fishr} %\footnotesize{(ICML'22\cite{rame2022fishr})} 
& 85.5 & 77.8 & 67.8 & 47.4 & \underline{41.7} & 64.0 \\
& %\cmark & \cmark
\cmark & \xmark
& MIRO$^{\dagger}$ \cite{cha2022domain}
%\footnotesize{(ECCV'22\cite{cha2022domain})} 
& 85.4 & \textbf{79.0} & {\bf 70.5} & \textbf{50.4} & \textbf{44.3} & \textbf{65.9} \\
\hline
%\multirow{9}{*}{Augmentations}
%& ERM$^{\dagger}$ &  84.2 & 77.3 & 67.6 & 47.8 & 44.0 & 64.2  \\
\parbox[t]{2mm}{\multirow{8}{*}{\rotatebox[origin=c]{90}{Augmentations}}}& 
%\xmark & \xmark
- & -
& ERM & 84.1 & 76.7 & 64.8 & 47.0 & 41.9 & 62.9 \\
& %\xmark & \xmark 
\xmark & \xmark
& + Mixup \cite{zhang2017mixup}  & 83.1 & 76.9 & 67.0 & 48.0 & 43.0 & 63.6  \\
%&& + I-Mixup$^{\dagger}$ & 84.6 & 77.4 & 68.1 & 47.9 & 39.2 & 63.4 \\
& %\xmark & \xmark 
\xmark & \xmark
& + CutMix \cite{yun2019cutmix} &  80.4 &  74.9 & 66.9  & \textbf{52.1} & 43.2 & 63.5 \\
& %\xmark & \xmark 
\cmark & \xmark
& + M-Mixup \cite{verma2019manifold}  &  84.2 &  77.8 &  67.4 & 46.3 & 43.0 & 63.7 \\
& %\cmark & \xmark 
\cmark & \cmark
& + MixStyle$^{\dagger}$ \cite{zhou2021domain} %\footnotesize{(ICLR'21\cite{zhou2021domain})} 
& 85.2 & 77.9 & 60.4 & 44.0 & 34.0 & 60.3 \\
& %\xmark & \xmark 
\xmark & \xmark
& +  BatchFormer \cite{hou2022batchformer} %\footnotesize{(CVPR'22\cite{hou2022batchformer})} 
& 82.7 & 76.7 & 65.8 & 48.6 & 42.8 & 63.3 \\
% & + Ours (1)$^{*}$ & 85.3 & 78.8 & 69.7 & 47.8 &   &  \\
% & + Ours (2)$^{*}$ & 85.9 & 79.2 & 67.1 & 48.1  &   &  \\
& %\xmark & \xmark 
\xmark & \xmark
& + \ours~ (SA) & \underline{85.8} \footnotesize{\color{cyan}{($\uparrow$1.7)}} & \textbf{78.7} \footnotesize{\color{cyan}{($\uparrow$2.0)}} & \textbf{69.0} \footnotesize{\color{cyan}{($\uparrow$4.2)}}  & \underline{49.9} \footnotesize{\color{cyan}{($\uparrow$2.9)}} & \textbf{45.1} \footnotesize{\color{cyan}{($\uparrow$3.2)}} & \textbf{65.7} \footnotesize{\color{cyan}{($\uparrow$2.8)}} \\
%& + Ours (SA)$^{*}$ & 85.8 & 78.7 & 69.0 & 49.8 & \textbf{45.1} & 65.7 \\
%& $\Delta$ & $\uparrow$ \color{cyan}{1.7} & - \color{cyan}{($\uparrow$2.0)} & - \color{cyan}{($\uparrow$4.2)} & - \color{cyan}{($\uparrow$2.8)} & - \color{cyan}{($\uparrow$3.2)} & - \color{cyan}{($\uparrow$2.8)} \\
& %\xmark & \xmark 
\xmark & \xmark 
& + \ours~ (Pool) & \textbf{86.3} \footnotesize{\color{cyan}{($\uparrow$2.2)}} & \underline{78.0} \footnotesize{\color{cyan}{($\uparrow$1.3)}} & \underline{68.4} \footnotesize{\color{cyan}{($\uparrow$3.6)}} & 49.5 \footnotesize{\color{cyan}{($\uparrow$2.5)}} & \underline{43.9} \footnotesize{\color{cyan}{($\uparrow$2.0)}} & \underline{65.2} \footnotesize{\color{cyan}{($\uparrow$2.3)}} \\
\hline\hline
%\multirow{8}{*}{Algorithms ++} 
% & Mixstyle$^{\dagger}$ & 85.2 & 77.9 & 60.4 & 44.0 & 34.0 & 60.3 \\
% & + BF &  &  &   &   &   &  \\
% & + Ours (2) &  &  &   &   &   &  \\
% & + Ours (3) &  &  &   &   &   &  \\
%\cline{2-8}
% & SWAD$^{\dagger}$ & 88.1 & 79.1 & 70.6 & 50.0 & 46.5 & 66.9 \\
% & + BF &  &  &   &   &   &  \\
% & + Ours (2) & 87.906 & 78.778 & 70.926 & 52.861  &   &  \\
% & + Ours (3) & 87.547 & 78.946 & 70.242 & 50.885  &   &  \\
%\hline\hline
\end{tabular}
\end{center}
\caption{{\em Results on domain generalization benchmarks.}~All models use a ResNet-50 backbone pre-trained on ImageNet. We use $\dagger$ to denote numbers that are reported from 
\cite{cha2021swad} and \cite{rame2022fishr}, while the rest are reproduced. See Appendix~\ref{appendix:break-down} for breakdown numbers and model variance. 
We categorize methods on the left into algorithms and augmentations, and use (1) to denote methods that need to be applied in many intermediate layers of the network; and use (2) to denote methods that require explicit domain information to guide learning.
We use cyan to denote the improvement over the robust ERM baseline \cite{gulrajani2020search}, and use bold and underline to highlight the highest number and the second highest number in both algorithms and augmentations sections.
Note that training MIRO requires guidance from pre-trained models.
}
\label{table-main}
\vspace{-3mm}
\end{table*}

\vspace{-2mm}
\subsubsection{Regularization with classifier guidance}
Thus, for each sample $x$, we created three latent and label pairs: $(z, y)$ as the original ones, $(z_d, \tilde{y})$ as the degraded ones, and $(z_r, y)$ as the restored ones. We perform a relaxed regularization with classifier guidance, where our model aims to optimize the below combined loss:
\[
\mathcal{L} =  \underbrace{\ell(g(z), y)}_{\text{original}} +  \underbrace{\ell(g(D_{\mu}(z), \tilde{y})}_{\text{degraded}} + \underbrace{ \ell(g(R_\theta(D_{\mu}(z)), y)}_{\text{restored}} 
\]
where $\ell$ is a classification loss function (e.g. CrossEntropy).

Crucially, our method relies on the classifier to both guide the corrupted latents towards degradation and back to their correct labels after restoration. This provides a soft constraint that encourages the latent embeddings to be flexible enough to mix information across domains.
% \zkn{The fact that mini-batches contain multi-domain samples should be emphasized in the text!}. 
Additionally, both operations use sample-to-sample relationships to create distribution-aware degraded and restored samples, which can further encourage the training set to be diversified.
\ours~fits perfectly into mini-batch training, as shown in pseudocode in Appendix Alg.~\ref{algo:main}.

\subsection{Assessing representation quality with alignment and uniformity metrics}

To assess the quality of representations learned through our model, we use two metrics shown to have good correlation with accuracy of classification from pre-trained models \cite{wang2020understanding}.  The first metric is the {\em alignment score}, a measure of the `closeness' of features from the same class. The second metric is the {\em uniformity score}, which measures the distribution of the (normalized) features on
the hypersphere and captures how well the representations span the space.

The alignment score is defined as:
\begin{equation}
\mathcal{L}_{\text {align }}(f) \triangleq \mathbb{E}_{(x_1, x_2) \sim p_{\text {cls }}}\left[\|f(x_1)-f(x_2)\|_2^2\right],
\end{equation}
where $(x_1, x_2) \sim p_{\text {cls}}$ draws samples from the same class.

The uniformity score is defined as:
\begin{equation}
\mathcal{L}_{\text {uniform }}(f) \triangleq \log \mathbb{E}_{x_1, x_2 \stackrel{\text { i.i.d. }}{\sim} p_{\text {data }}}\left[e^{-2\|f(x_1)-f(x_2)\|_2^2}\right],
\end{equation}
where in this case the samples $(x_1, x_2)$ are across the entire dataset. Uniformity demonstrates the generalization ability of the method, as it measures how much information the model preserves from the training data.

%%%%% RESULTS
\section{Experimental Results}
To examine the performance of \ours~on diverse datasets, we study our approach on:
% two domain generalization tasks: 
(i) domain generalization benchmarks, and (ii) medical imaging datasets with strong domain shifting.
To further demonstrate its potential in different applications, we also test our method on (iii) long-tail recognition with imbalanced classes.

\subsection{Domain generalization}

\subsubsection{Experiment setup}

\paragraph{Dataset.}
Following \cite{gulrajani2020search}, we comprehensively evaluate our method on the standardized DomainBed benchmarks. Specifically, five datasets are used in our experiments:
(1) \underline{PACS} \cite{li2017deeper} (9,991 images, 7 classes, and 4 domains), (2) \underline{VLCS} \cite{fang2013unbiased} (10,729 images, 5 classes, and 4 domains), (3) \underline{Office-Home} \cite{venkateswara2017deep} (15,588 images, 65 classes, and 4 domains), (4) \underline{TerraIncognita} \cite{beery2018recognition} (24,788 images, 10 classes, and 4 domains), (5) \underline{DomainNet} \cite{peng2019moment} (586,575 images, 345 classes, and 6 domains).

% Above information are copied from SWAD \cite{cha2021swad}. Is it okay to just write it this way? or should we alter text?

\vspace{-3mm}
\paragraph{Evaluation and experimental details.} For a fair comparison, we follow the leave-one-domain-out model training and evaluation protocol as in prior works \cite{cha2021swad,cha2022domain}.
% where 20$\%$ samples are used for model selection. 
% Specifically, for each dataset, each domain is selected once as the target domain, while the remaining domains are used to train the model where 20$\%$ samples are used for validation and model selection.
Unless specified otherwise, the same dataset splits (80/20$\%$), hyperparameter sets, and model optimizer in \cite{cha2021swad} are used, % to reduce computational costs, 
and a ResNet-50 pre-trained on ImageNet is used for weight initialization.
% and is optimized by an Adam optimizer.
For each training step, we construct a mini-batch containing 32 images from each training domain. For \ours, we perform a fixed learning rate adjustment due to our high loss value. We performed hyperparameter search on the validation sets of PACS and used the fixed hyperparameter sets on other experiments. All experiments are repeated five times with different random seeds except for DomainNet which has six subdomains.
%due to its computational cost.
More experimental details can be found in Appendix~\ref{appendix:alg-details}.

\vspace{-2mm}
\subsubsection{Results on DomainBed}
\vspace{-2mm}
In Table~\ref{table-main}, we evaluate our approach against a diverse set of algorithms that include general-purpose data augmentation techniques such as Mixup \cite{zhang2017mixup} and Manifold-Mixup \cite{verma2019manifold}, domain-specific data augmentation techniques like MixStyle \cite{hong2021stylemix}, 
%and RSC \cite{huang2020self}, 
as well as other DG algorithms that leverage domain information to build domain-invariant representations \cite{sun2016deep,rame2022fishr}.
%such as CORAL  and Fishr. 
On the left, we provide a breakdown of model assumptions used in the different models tested to highlight the flexibility of our approach and limited use of additional assumptions.

Our approach outperforms the robust ERM baseline \cite{gulrajani2020search} and shows significantly more stable and robust performance compared to other augmentation methods like BatchFormer \cite{hou2022batchformer} and MixStyle \cite{zhou2021domain}, resulting in a 2.7\% improvement over ERM when averaged over all 5 datasets in DomainBed and a $\approx$2.0\% improvement over other augmentation methods. Our approach shows particularly robust performance on DomainNet, the largest DG dataset with the most classes, highlighting the scalability of our approach. 
% Our approach beats all of the other methods on DomainNet, the largest dataset with the most classes, highlighting the scalability of our approach. 
%Overall, we achieve performance on par with the state-of-the-art method MIRO that regularizes features at both early and later layers in a pre-trained encoder to filter out features in early layers that are not beneficial for transfer. 
Overall, we achieve performance on par with the state-of-the-art method MIRO that regularizes features at many intermediate layers of the encoder, and requires an unaltered pre-trained encoder to provide guidance throughout training. 
The breakdown of accuracies for each dataset reported across domains and model variance across random seeds are in Appendix~\ref{appendix:break-down}.

%Our approach effectively degrades and restores the latent representations of the samples, leading to improved generalization performance. Furthermore, our approach does not require any domain-specific knowledge or additional training data, making it highly efficient and scalable for large-scale applications.

\vspace{-2mm}
\subsubsection{Ablations and visualizations}
\vspace{-2mm}
To further understand the robustness of the proposed method, we perform model ablation, latent representation evaluation, and nearest neighbor visualization
with our model on the PACS dataset.

\vspace{-3mm}
\paragraph{Ablations.}~
\label{para:degrade}
Our approach consists of two main components, a step to degrade samples and a step to restore them. Thus, we wanted to understand how both steps contribute towards the overall performance.
In Table~\ref{table-ablation}, we compare the performance of our method when we use both D+R vs. when we only apply degradation (D-only) or restoration (R-only). Across all tested domains in PACS, we see that the combination of both D+R is consistently the best. However, when we remove restoration and use D-only, we also obtain good performance, suggesting that sample-aware degradation can also be a good augmentation on its own. \footnote{If there is no degradation, the cross-attention operation becomes a self-attention operation, and thus our method converges back to \cite{hou2022batchformer}.}

%{\bf Testing Gaussian noise as a source of degradation.}~~
To test whether a simpler degradation mechanism could be used, we tested a variant of our model where we remove the sample-aware mixing operation used in our degradation step and replace it with i.i.d. Gaussian noise $\mathcal{N}(0, 1)$.  When no restoration is used (D-only), we find that the Gaussian noise augmentation does not help; However, when we couple Gaussian noise with our restoration procedure (R-only, D+R), we find that we can obtain good performance. Overall, we observe that combining both degradation and restoration achieves the best overall performance, and our sample-aware approach provides a flexible and robust way to augment latents.

%Without any latent degradation, our method boils down to \cite{hou2022batchformer}, which does not systematically improve performance in many benchmarks. However, when replacing sample-aware latent degradation with additive Gaussian noise, latent restoration (Gaussian R-only) outperforms the ERM baseline, indicating that our restoration procedure could be effective when combined with other forms of  degradation.

%This suggests that our proposed latent degradation and restoration procedures work in tandem, and it is critical to consider sample-to-sample relationships to have the most impact on out-of-domain generalization. 

%To understand whether the degradation is indeed producing outputs that confuse the classifier, we examined the classification accuracy for the degraded latents and obtained an average accuracy of $33\%$ for a 7-way classification (random guess rate is $14\%$). 

\begin{table}[t]
\begin{center}
% \begin{small}
% \begin{sc}
\begin{tabular}{lcccccr}
%\hline
&  & A & C & P & S & Avg. \\
\hline
Reference & ERM & 84.9 & 80.6 & 95.9 & 75.0 & 84.1 \\
\hline
\multirow{3}{*}{Ours} 
& D-only & 85.7 & 78.6 & 96.9 & 78.2 & \underline{84.8} \\
& R-only & 82.6 & 79.3 & 95.4 & 73.5 & 82.7 \\
& D+R & 86.3 & 82.6 & 97.1 & 79.2 & \textbf{86.3} \\
\hline
\multirow{3}{*}{Gaussian} 
& D-only & 85.0 & 78.1 & 96.5 & 75.6 & 83.8 \\
& R-only & 86.4 & 80.0 & 97.4 & 76.5 & \underline{85.1} \\
& D+R & 87.4 & 79.7 & 97.1 & 77.1 & \textbf{85.3} \\
\hline
\end{tabular}
% \vskip 0.15in
% \end{sc}
% \end{small}
\end{center}
\caption{\textit{Ablations of degradation and restoration on PACS}. We compare the performance of \ours~using the degradation loss (D-only) or restoration loss (R-only) alone vs. combining them (D+R), for our sample-aware degradation approach (top) vs. additive Gaussian noise (below).}
\label{table-ablation}
\vspace{-3mm}
% \vskip -0.1in
\end{table}

\begin{figure*}[t!]
\begin{center}
 \includegraphics[width=\textwidth]{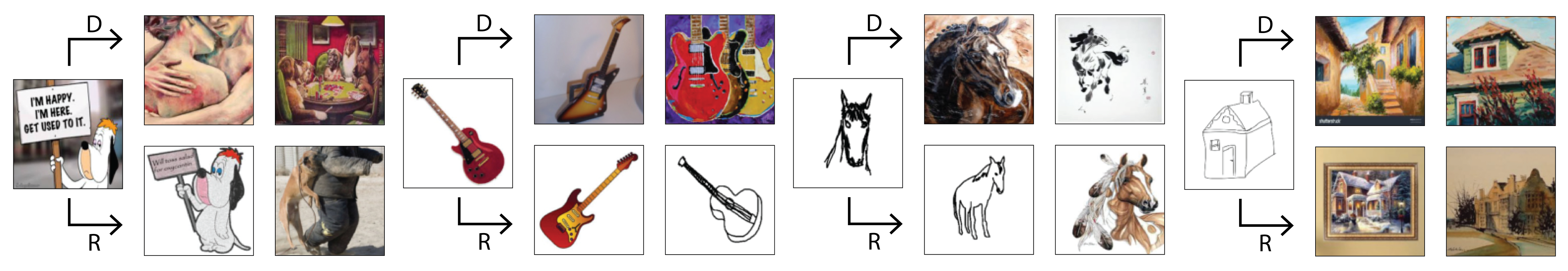}
 \end{center}
\vspace{-2mm}
\caption{
\small {\em Nearest neighbors visualizations.} We show the top two nearest neighbors for a query sample after degradation (top row) and restoration (bottom row). In the leftmost example, we find a case where degradation places a sample near points from a different class but brings it back to diverse domains from its original class. On the right, we show examples where degradation maps the latent next to samples of the same class that are different in color or background, and restores to samples that are similar but might be from another domain. 
%Latent restoration transports samples into different parts of the latent space
\label{fig:neighbors}}
%\vspace{-4mm}
\end{figure*}

 \begin{figure}[t!]
\begin{center}
 \includegraphics[width=0.46\textwidth]{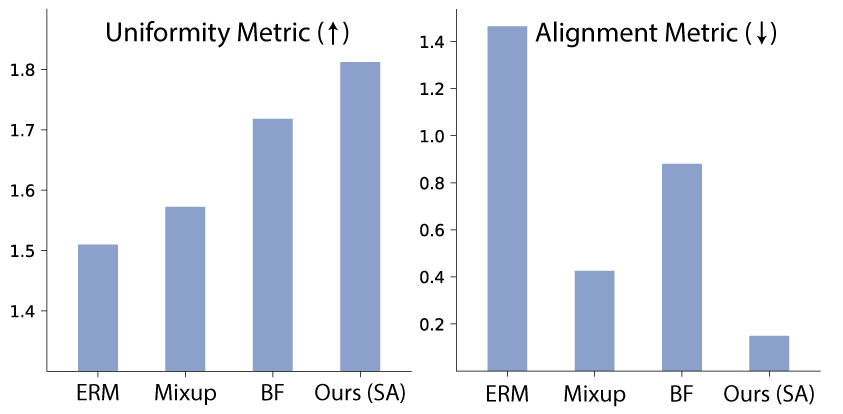}
 \end{center}
   \caption{\textit{Latent quality evaluation.} We measure the uniformity and alignment of the latent spaces for ERM, Mixup \cite{zhang2017mixup}, BatchFormer \cite{hou2022batchformer} (BF), and \ours~(Ours-SA).\vspace{-2mm}}
 \label{fig:latents-metrics}
 \vspace{-2mm}
\end{figure}

%\vspace{-3mm}
%\paragraph{Sharing classifier weights.}
% Our model trains a classifier $g$ on top of the encoder $f$, and uses the same classifier $g$ to regularize the model to learn degraded samples and restored samples. Following \cite{hou2022batchformer}, we tested whether training two different classifiers, one for the original loss, and the other for the corruption/restoration steps, would benefit the training. With a separate classifier, our model gets an average of $83.7\%$ accuracy on PACS, an over 2\% decrease. It seems it is critical to share classifier weights for our method.
% , different from \cite{hou2022batchformer}.

\vspace{-3mm}
% \paragraph{Examining the latents and the effect of restoration.} 
\paragraph{Latent quality evaluation.} 
To understand how our proposed approach shapes the representations of the data, we examined the uniformity and alignment metrics for different models as shown in Figure~\ref{fig:latents-metrics} (see Appendix~\ref{appendix:latents} for direct visualizations). We confirmed that our method  gives both the highest uniformity score, which demonstrates that it preserves the maximum amount of information inside the training data  and the lowest alignment score, which demonstrates that it encourages the closeness of latents within the same class. Interestingly, we find that Mixup \cite{zhang2017mixup} does improve alignment but has very little improvement in uniformity when compared with ERM. On the other hand, BatchFormer \cite{hou2022batchformer} produces good diversity of latents by considering nonconvex sample-to-sample relationships but poorer alignment. Our method, instead, combines the advantages at both ends by encouraging the diversity of the generated samples while maintaining good alignment by separating the different classes.

\vspace{-3mm}
\paragraph{Nearest neighbor visualization.} 
To observe how the degradation and restoration steps remap different query points, we examined the nearest neighbors of the augmented latents for both operations (Figure~\ref{fig:neighbors}). We observed that indeed the latents produced from the latent restoration process would retrieve nearest neighbors to samples from the same class but different domains (both the testing domain and another training domain).
% near samples that are from the same class, but often the new nearby point is from a different domain (both the testing domain and another training domain). 
Interestingly, in the degradation step, although the produced samples would often confuse the classifier, the nearest neighbors are often from the same class but with dissimilar colors or backgrounds.
% the class may be switched but more often than not when the model is fully trained, although the nearest neighbors typically would match samples with different colors or backgrounds.
We conjecture that it is possible that this classifier guidance may encourage certain content/style separation in the latent space.

\subsection{Experiments on medical imaging datasets}

To further demonstrate the use of our method across different tasks and images, we applied the method to multiple medical imaging datasets where domain shift occurs and generalization is difficult due to small sample sizes. 

\vspace{-2mm}
\subsubsection{Experiment setup}
\vspace{-1mm}
\paragraph{Datasets.} We evaluate our method on five medical imaging datasets: \underline{Derma} \cite{tschandl2018ham10000} (10,015 samples, 7 classes, 2D), \underline{OrganS} \cite{bilic2023liver} (25,221 samples, 11 classes, 2D), \underline{OCT} \cite{kermany2018identifying} (109,309 samples, 4 classes, 2D), \underline{Fracture} \cite{jin2020deep} (1,370 samples, 3 classes, 3D), as well as \underline{Camelyon17} \cite{bandi2018detection} (302,436 samples, 2 classes, 2D).
For the first four datasets, we follow the preprocessing and data splitting protocol as in \cite{yang2023medmnist}; while for Camelyon17 we followed % a domain shift benchmark called Wilds
\cite{koh2021wilds}.
More details about datasets and their domain shifts are in Appendix~\ref{appendix:alg-details}.

\begin{table}[b]
\vspace{-2mm}
\vskip 0.15in
\begin{center}
% \begin{small}
% \begin{sc}
\setlength\tabcolsep{4pt}
\resizebox{\columnwidth}{!}{%
\begin{tabular}{l cccc r}
% \hline
Model & Derma & OrganS & OCT & Fracture & Avg. \\
% \hline
% classes & 7 & 11 & 4 & 3 &  \\
% samples & 10,015 & 25,221 & 109,309 & 1,370 &  \\
% dimension & 2D & 2D & 2D & 3D &  \\
\hline
ERM$^{\dagger}$ & 75.4 & 77.8 & 76.3 & 50.8 & 70.1 \\
ERM & 75.8 & 79.8 & 76.4 & 49.6 & 70.4 \\
% + Aug &  &  &  &  &  \\
+ Mixup & 70.0 & 80.8 & 78.9 & 48.3 & 69.5 \\
% \hline
+ M-Mixup & 76.4 & 80.1 & 76.3 & 49.6 & 70.6 \\
+ BF & 76.8 & 82.7 & 77.8 & 51.7 & 72.3 \\
\hline
+ Ours (SA) & 78.1 & {\bf 83.3} & 79.1 & 52.1 & 73.2 \\
+ Ours (Pool) & {\bf 78.4} & 82.9 & {\bf 79.4} & {\bf 55.0} & {\bf 73.9} \\
\hline
\end{tabular}
}
\setlength\tabcolsep{6pt}
% \end{sc}
% \end{small}
\end{center}
\caption{\textit{Medical image classification without explicit domains}. We follow the setup in \cite{yang2023medmnist}, where $^{\dagger}$ are the classification accuracies reported by them with the same ResNet-18 backbone model.\label{table:medmnist}
% \colortext{Note that the variance is slightly larger. Do repeat the experiments and take the mean}
}
\end{table}

\vspace{-3mm}
\paragraph{Evaluation and experimental details.}
For the first four datasets, we follow 
% the training pipeline and dataset splits in
\cite{yang2023medmnist} and use a ResNet-18 for 2D image classification and a ResNet-18-based 3D backbone for 3D image classification. We replace the Adam optimizer with an SGD optimizer with lr=$0.001$ and momentum=$0.9$ for better performance. %\zkn{tuned how?}.
% to benchmark the performance. 
% We replaced the Adam optimizer as used in \cite{yang2023medmnist} with an SGD optimizer with lr=$0.001$ and momentum=$0.9$, and batch size=$128$. All models are trained for 100 epochs till convergence. 
For Camelyon17, we follow the model and training setup in \cite{shi2021gradient} and use a DenseNet-121 backbone.
% , an SGD optimizer with lr=$0.0001$ and momentum=$0.9$, and batch size=$32$. All models are trained for 20 epochs. 
%The hyperparameters and optimizers remain the same as previous methods. 
More information about the training and model optimization process is in Appendix~\ref{appendix:alg-details}.
For \ours, we performed the same learning rate adjustment, and applied the same set of hyperparameters as defined in previous experiments, demonstrating the stability our method across various choices of hyperparameters.
% demonstrating insensitivity to hyperparameter tuning\zkn{Not clear how competing methods are tuned (or not). }\zkn{Added emphasis on the fact that no tuning shows it's relatively insensitive to hyper-parameters}. 
All experiments are repeated three times with different random seeds.

\begin{table}[t]
% \vspace{-2mm}
\begin{center}
% \begin{small}
% \begin{sc}
\begin{tabular}{lccr}
% \hline
Model & Without Aug & Targeted Aug & Avg. \\
\hline
CORAL$^{\dagger}$ & 59.5 & $-$ &  \\
Fish \cite{shi2021gradient} & 74.7$^{\dagger}$ & 76.9 & 75.8 \\
ERM$^{\dagger}$ & 70.8 & 82.0 & 76.4 \\
ERM & 72.3 & 84.3 & 78.3 \\
+ Mixup$^{\dagger}$ & 63.5 & $-$ &  \\
% + M-Mixup & & &  \\
% + BF & & &  \\
\hline
+ Ours (SA) & 76.4 & \textbf{91.0} & 83.7 \\
+ Ours (Pool) & \textbf{81.3} & 89.9 & \textbf{85.6} \\
\hline
\end{tabular}
% \end{sc}
% \end{small}
\end{center}
\caption{\textit{Medical image classification on Camelyon17}. We follow the setup in \cite{shi2021gradient}, where $^{\dagger}$ are the classification accuracies reported in \cite{shi2021gradient,koh2021wilds} with the same DenseNet-121 backbone model.
%\zk{Is DenseNet-121 the ERM model? It's confusing to have the methods (Fish/CORAL) above them, as it makese it seem like the other ones use something other than DenseNet-121. Perhaps also add ``(ERM)'' to clarify, or just use ERM like Table 1. 
\label{table:camelyon}\vspace{-2mm}}
% \vskip -0.1in
\end{table}

\vspace{-3mm}
\subsubsection{Results: Medical imaging datasets}
\vspace{-2mm}
In Table~\ref{table:medmnist} and \ref{table:camelyon}, we compare our model with other baseline and competitor models. 
For the first four datasets where we do not have explicitly defined different domains, we benchmark our method with other augmentation methods \cite{zhang2017mixup,verma2019manifold,hou2022batchformer}. We find that augmentation on image space (Mixup) does not always improve performance,
as blending the color or the shape of medical objects might undermine the classification of specific features. 
In terms of latent augmentation methods, we observed that BatchFormer \cite{hou2022batchformer}, which considers sample-to-sample relationships, also has good performance. 
However, our method outperforms all other methods, demonstrating the effectiveness of considering both sample-to-sample relationships and classifier-guided latent degradation.

When testing on Camelyon17, we compare models under two different settings: (i) standard evaluation setting without custom augmentations \cite{koh2021wilds}, and (ii) with targeted augmentations that use underlying knowledge about the type of domain shift in the dataset.  
In the first setting, our model outperforms many reported algorithms by a large margin and improves over the ERM baseline by nearly 10\% (from 72.3\% to 81.3\%), where the performance of our method almost reaches the performance of ERM with targeted augmentation (84.3\%). When testing on targeted augmentations, we find an impressive boost of almost 7\% over the ERM baseline that is competitive with state-of-the-art for this task. In contrast, Fish \cite{shi2021gradient} remains at around 75\% in both the original case and in targeted augmentations. %, suggesting that the gradient-based regularization may somehow stifle improvements obtained through . % \zkn{Did you tune the algorithm's hyper-parameters?}\zk{Is Fish the current state of art for these datasets? Might want to mention, for each dataset, which ones are state of art. Otherwise it's hard to understand why you're comparing to them, different from the first DG table}. %and does not perform well. 
This application shows the great potential of \ours~when applied to medical imaging datasets where data augmentation requires domain knowledge to define.

\subsection{Application to long-tail recognition}

%Inspired by the high performance of using sample-to-sample relationships \cite{hou2022batchformer} on long-tail recognition tasks, 
% benefit the classification when facing sample imbalance issue,
Augmentation methods are shown to be effective in improving model generalization when datasets exhibit class imbalance \cite{li2021metasaug,liu2020deep}.
To examine if \ours~can also enhance generalization in the presence of strong data imbalance,
%\zkn{Might be good to motivate why this makes sense, e.g. how data augmentation can help this imbalance}, 
we applied our method to a long-tail recognition task.% to evaluate the effectiveness of our approach under a different scenario.

\vspace{-3mm}
% \subsubsection{Experiment setup}
\paragraph{Experiment setup.} 
We perform our experiments on CIFAR-100-LT \cite{krizhevsky2009learning} with an imbalance ratio of 100. 
\ours~is applied
% We applied our method 
on top of BalancedSoftmax (BALMS \cite{ren2020balanced}) and Balanced contrastive learning (BCL \cite{zhu2022balanced}), which both have explicit CrossEntropy term in their loss functions. We followed their hyperparameter and training settings, and for our method we performed the same learning rate adjustment as in domain generalization experiments. More information is in Appendix~\ref{appendix:alg-details}.

\vspace{-3mm}
%\subsubsection{Results: Long-tail recognition}
%\vspace{-2mm}
\paragraph{Results.}
As shown in Table~\ref{table:longtail},
our method surpasses both the original method and \cite{hou2022batchformer}, demonstrating its potential for being a versatile plugin module. Interestingly, different from \cite{hou2022batchformer} that argues sample-to-sample relationship mostly improves performance on tail classes, our method gains the largest performance improvements on Medium classes, which might be attributed to our soft-label construction process.
% It might due to the fact how the sample-to-sample relationship is constructed. 
Additionally, we observed that our method performs better when combined with a simpler method (BALMS).
% Also, our method seems to perform better when the method is relatively simpler (BALMS). 

\begin{table}[t]
\begin{center}
% \begin{small}
% \begin{sc}
\begin{tabular}{lcccr}
% \hline
% Many Medium Few
& Many & Medium & Few & Avg. \\
\hline
BCL \cite{zhu2022balanced} & 69.1 & 52.4 & \textbf{30.5} & 51.7 \\
 + BF & \textbf{69.5} & 51.7 & 27.9 &  50.8 \\
 + Ours & 68.9 & \textbf{53.9} & \textbf{30.5} & \textbf{52.1} \\
 \hline
BALMS \cite{ren2020balanced} & 67.9 & 50.7 & 32.2 & 51.1 \\
 %+ BF & 68.8  & 50.6 & 31.2 & 51.2  \\
 % 51.7 68.4 49.3 34.3 :: this is what is reported in BF paper
 + BF$^\dagger$ & 68.4  & 49.3 & \textbf{34.3} & 51.7  \\
 + Ours & \textbf{69.8} & \textbf{51.3} & 34.2 & \textbf{52.7}  \\
\hline
\end{tabular}
% \end{sc}
% \end{small}
\end{center}
\caption{\textit{Long-tail recognition results on CIFAR-100-LT}. We reproduced all numbers except for BALMS+BF, where we used their reported numbers \cite{hou2022batchformer} as it is higher than what we reproduced.}
\label{table:longtail}
\end{table}

%%%%% CONCLUSION

%%%% CONCLUSION
\section{Conclusion}

In this paper, we proposed a novel latent augmentation method \ours~to improve domain generalization. The idea behind our approach is to learn to  {\em degrade} and {\em restore} latents at the batch level, using information across 
%the entire batch 
samples to build augmentations for a latent. We use a classifier to guide both steps: first to obtain degraded latents that are identified 
% as the original class label 
to a constructed soft-label
and second, to get the latents restored to the original class.
% Importantly, t
The proposed method can be easily integrated into existing deep learning methods and used with different encoders without any modifications, achieving significant improvements.

This work opens up a number of interesting future directions including the integration of our method with domain adaptation approaches, as well as exploring its use 
% with other data augmentation techniques, and
in semi-supervised learning and self-supervised learning. It would be interesting to study the connections between our method and generative modeling frameworks that use cycle consistency and cold diffusion \cite{bansal2022cold}.

%%%% CONCLUSION
\section{Acknowledgement}

This project is supported by NIH award 1R01EB029852-01, NSF award  IIS-2039741, NSF award IIS-2212182, NSF CAREER award IIS-2146072, the NSF Graduate Research Fellowship Program (GRFP) for MD, as well as generous gifts from the Alfred Sloan Foundation, the McKnight Foundation, the CIFAR Azrieli Global Scholars Program.

%\vspace{-3mm}
%\paragraph{Limitations and Future Work.}~However, our work is not without limitations. For example, our method is only evaluated on a limited number of tasks, and its effectiveness in other domains is yet to be explored. Moreover, our current approach is based on cross-attention -- exploring other translation-based methods and other transformer architectures could be an interesting avenue for future research. Future work can investigate the integration of our approach with domain adaptation approaches,  exploring its use with other data augmentation techniques, and in self-supervised learning. It is also worth investigating the theoretical insights, especially how the proposed method could be related to generative modeling frameworks that use cycle consistency and diffusion.
%Moreover, similar as \cite{cha2022domain,robey2021model}, we used additional parameters during training, which varies based on architecture selection. Although they are removable during inference, moving forward, it would be interesting to investigate how to reduce/remove this parameter constraint. 

{\small
\bibliographystyle{ieee_fullname}
\bibliography{main}
}

\newpage
\renewcommand{\thesubsection}{\Alph{subsection}}

\clearpage
\section*{Appendix}
\setcounter{page}{1}

\subsection{Algorithm and implementation details}
\label{appendix:alg-details}

\subsubsection{Additional motivation and hypothesis}

\paragraph{The degradation-restoration process}
One critical aspect of this work is that the degradation and restoration processes are optimized {\textbf{jointly}} {\textbf{instead of individually}}.
One hypothesis of ours is that \ours~implements an \textit{implicit adversarial process}: The degradation step $D$ aims to extract features that are (incorrectly) considered as
non-discriminative
% considered non-discriminative
by the classifier $g$, while the restoration step $R$ learns to recover discriminative components (which might be otherwise ignored due to overfitting).
As shown in Table 2 in the paper, our ablations demonstrate that only when $D$ and $R$ are jointly applied, is there a significant enhancement in performance.
Additionally, our restoration step $R$ is {\textbf{guided by the classifier}}, which prevents it from being trivial in contrast to a Euclidean distance. We provide more evidence through visualizations as shown in Figure~\ref{fig:all}(A), the majority of the restored samples ({\color{darkergreen}{$\blacktriangledown$}}) are shifted away from their queries ({\color{red}{$\star$}}), and many are out of the training distribution (gray).

\paragraph{Distribution awareness}

Another crucial aspect of this work is the use of distribution awareness. In our ablation experiments, we show that using a distribution-aware approach to create degraded samples {\textbf{outperforms unstructured perturbations}} generated through additive Gaussian noise.
We hypothesize that since \ours~apply $D$ and $R$ as transformers that consider sample-to-sample relationships,
they would alter the gradient flow \cite{hou2022batchformer}, and help the encoder $f$ to learn better representations.
% R2 asked about what {\color{purple}{`assumptions of distribution'}} we need for the approach to work. 
While 
our experiments and ablations provide strong evidence that suggests that LatentDR can be used as a plug-and-play approach in diverse tasks (domain generalization, long-tail recognition, and medical imaging classification), without strong assumptions or modifications of the sample/label distributions, it remains an interesting problem on which assumption of distribution \ours~need for it to work.

\subsubsection{Algorithm details}

\paragraph{Pseudocode}
\ours~can be implemented through the below pseudocode. During the training stage, the degradation operator $D$ and the restoration operator $R$ would create latent augmentations $(z_d, \tilde{y})$ and $(z_r, y)$, correspondingly. The augmented latents would guide the model to learn the relationship across samples, and thus generalize better to unseen sources  which span several training domains. During inference, both the  degradation and restoration operators are removed.
\begin{algorithm}[h]
\label{code:alg}
\SetAlgoLined
    % \PyComment{going to have indentation} \\
    \PyCode{def train(z, g, y, D, R):} \\
    \Indp   % start indent
        \PyComment{z: a batch of latents} \\
        \PyComment{g: a classifier} \\
        \PyComment{y: one-hot labels of z} \\
        \PyComment{D: the degradation operator} \\
        \PyComment{R: the restoration operator} \\
        % \PyComment{y: label} \\
        \PyCode{z$_d$ = D(z)} \\ % \PyComment{inline comment} \\ 
        \PyCode{z$_r$ = R(z$_d$, z)} \\
        \PyCode{$\tilde{\text{y}}$ = y.sum(0)/y.shape[0]} \\
        \PyCode{l1 = F.cross$\_$entropy(g(z), y)} \\
        \PyCode{l2 = F.cross$\_$entropy(g(z$_d$), $\tilde{\text{y}}$)} \\
        \PyCode{l3 = F.cross$\_$entropy(g(z$_r$), y)} \\
        \PyCode{loss = l1 + l2 + l3} \\   
        \PyCode{loss.backward()} \\
    \Indm % end indent, must end with this, else all the below text will be indented
    - \\
    \PyCode{def pred(z, g):} \\
    \Indp
        \PyComment{The prediction stage} \\
        \PyCode{return g(z)} \\
    \Indm
%    \PyComment{this is a comment} \\
%    \PyCode{your code}
\caption{Pseudocode of \ours}
\label{algo:main}
\end{algorithm}

\paragraph{Algorithm variants}

For the degradation and restoration operators $D$ and $R$, we tested two simple variants of the transformer layer. The formulation of the two variants are denoted as below:
\begin{equation}
\label{eq:msa-1}
\begin{aligned}
    & Z_{\ell+1}^{\prime} = \operatorname{LN}(Z_{\ell} + \operatorname{AttN}(Z_{\ell})) \\
    & Z_{\ell+1} = \operatorname{LN}(Z_{\ell+1}^{\prime} + \operatorname{FF}(Z_{\ell+1}^{\prime})), ~0 \leq \ell \leq L-1
\end{aligned}
\end{equation}
\begin{equation}
\label{eq:msa-2}
\begin{aligned}
    & Z_{\ell+1}^{\prime} = \operatorname{LN}(Z_{\ell}) + \operatorname{AttN}(Z_{\ell}) \\
    & Z_{\ell+1} = \operatorname{LN}(Z_{\ell+1}^{\prime}) + \operatorname{FF}(Z_{\ell+1}^{\prime}), ~0 \leq \ell \leq L-1
\end{aligned}
\end{equation}
where AttN denotes either the self-attention operation (MSA($\cdot$)) or the pooling operation (Pool($\Omega(\cdot)$)) for \ours~(SA) or \ours~(Pool). In equation~\ref{eq:msa-1} and \ref{eq:msa-2}, the layerwise normalization (LN) is applied later than or prior to the attention (AttN) and feedforward (FF) operations, respectively.
We used the equation~\ref{eq:msa-1} variant for our experiments in DomainBed and medical imaging classification tasks, where we did not observe a significant difference between the two variants. In long-tail recognition experiments, we used the equation~\ref{eq:msa-2} variant, where we observed that it outperforms the other variant by a large margin.

% we use one layer Transformer Encoder for both the degradation process and the restoration process.

\subsubsection{Implementation details}

In all our experiments, we used a one-layer Transformer encoder for both the degradation process and the restoration process for simplicity.

\paragraph{DomainBed experiments}

For all of the benchmark models, we follow the training and evaluation protocol as in \cite{cha2021swad}, where we used their default algorithm-agnostic hyperparameters including batch size, learning rate, dropout, and weight decay. Due to the large loss value of \ours, we performed a fixed learning rate adjustment of $50\%$ for all of our experiments, and fixed the other hyperparameters. \ours~also contains many algorithm-specific parameters, including the dimensionality of the transformers (both the attention head dimension $\operatorname{dim-head}$ and the feed-forward dimension $\operatorname{dim-ff}$), and transformer dropout rate. %, and the weights for different loss terms as below:
Searching all hyperparameters on each dataset would require heavy computational resources. Thus, we search the  algorithm-specific hyperparameters on the PACS datasets, and use the same hyperparameters on all other datasets.

For \ours~(SA), we search $\operatorname{dim-head}$ and $\operatorname{dim-ff}$ in $[\operatorname{dim}, \operatorname{dim/2}, \operatorname{dim/4}, \operatorname{dim/8}]$, where $\operatorname{dim}$ is the latent space dimensionality. The transformer dropout rate is searched in $20\%, 50\%, 70\%$. We used $\operatorname{dim/4}$ for both dimensions, and used $50\%$ dropout rate after the parameter selection. Our competitor model BatchFormer \cite{hou2022batchformer} 
% requires the same hyperparameter searching process, and
is defaulted to select $\operatorname{dim}$ for both dimensions and a dropout rate of $50\%$ as in their paper. To ensure the fairness of comparison, we performed the same hyperparameter search for BatchFormer and used their default values as it returns the best results.

% and used the same dimensionality for  to ensure the fairness of comparison. % In practice, we observed that sharing the degradation and restoration operator weights in \ours~(SA) does not hurt the performance, as the self-attention operation can be treated as a special case of cross-attention operation where the condition vectors are in the same set of query vectors. Thus, we used the same weights to reduce additional parameter searching and training parameters. 

For \ours~(Pool), we search the hyperparameter $\operatorname{dim-head}$ in $[\operatorname{dim/8}, \operatorname{dim/16}, \operatorname{dim/32}]$ and the hyperparameter $\operatorname{dim-ff}$ in $[\operatorname{dim/4}, \operatorname{dim/8}]$. The dropout rate is fixed to be $50\%$ due to the search results in the previous model to reduce additional computational costs.
We used $\operatorname{dim-head}=\operatorname{dim/32}$, $\operatorname{dim-ff}=\operatorname{dim/8}$ after the parameter selection.
% As the Pooling operation is not equivalent to the cross-attention 
Interestingly, we noticed that replacing the self-attention operator with a pooling operator requires fewer additional parameters in training, while the additional stochasticity ensures the robustness of the performance. 

% , the parameter costs can be greatly reduced, while 

\paragraph{Medical imaging experiments}

For the first four datasets, we follow the training pipeline and dataset splits in \cite{yang2023medmnist}, where we used a ResNet-18 for 2D images and a ResNet-18-based 3D backbone to benchmark the performance.
The ResNet-18-based 3D backbone is built with the ACSConv package.
% \cite{yang2021reinventing}.
% We replaced the Adam optimizer as used in \cite{yang2023medmnist} with an SGD optimizer with lr=$0.001$ and momentum=$0.9$, and batch size=$128$. 
All models are trained for 100 epochs till convergence. 
For Camelyon17, we followed the model, data, and training setup in \cite{shi2021gradient} and used a DenseNet-121 backbone.
We used an SGD optimizer with lr=$0.0001$ and momentum=$0.9$. All models are trained for 20 epochs with a batch size=$32$.
%The hyperparameters and optimizers remain the same as previous methods. 
We applied the same set of hyperparameters based on our experiments in DomainBed, which demonstrated the robustness of our approach as it is insensitive to the selection of hyperparameters.

\paragraph{Long-tail recognition experiments}

Inspired by the robust performance of \cite{hou2022batchformer} on long-tail recognition tasks, we applied our method on CIFAR-100-LT. We add \ours~(SA) on top of BCL and BALMS following their default hyperparameter and training settings. For our model, we used the same hyperparameters as above, 
% We tested only \ours~(SA)
and used the eq.\ref{eq:msa-2} variant of transformer formulation to achieve the best performance in long-tail recognition tasks.

\subsection{Full results on DomainBed}
\label{appendix:break-down}

We provide the full results for all augmentation methods in all tested DomainBed datasets as below. Note that the details of the break-down performance for other algorithms can be found in \cite{cha2021swad} and \cite{rame2022fishr}.

\paragraph{PACS}
The breakdown performances are shown in Table \ref{appendix:bd_pacs}, where $\sigma$ is the standard deviation (SD).

\begin{table}[h]
\begin{center}
\label{table-pacs}
\vskip 0.15in

\begin{small}
\centering
% \begin{sc}
\begin{tabular}{lcccccr}
\hline
 & A & C & P & S & Avg. \\
\hline
ERM & 84.9 & 80.6 & 95.9 & 75.0 & 84.1 \\
\hline
+ Mixup & 84.2 & 79.4 & 96.0 & 72.8 & 83.1 \\
% + I-Mixup$^{\dagger}$ &  &  &  &  & 84.6 \\
+ CutMix & 81.5 & 75.8 & 95.6 & 68.9 & 80.4 \\
+ Manifold-Mixup & 84.9 & 79.5 & 96.7 & 75.8 & 84.2 \\
+ MixStyle$^{\dagger}$ & 86.8 & 79.0 & 96.6 & 78.5 & 85.2 \\
+ BatchFormer & 82.6 & 79.3 & 95.4 & 73.5 & 82.7 \\
\hline
+ LatentDR (SA) & 87.4 & 81.2 & 97.8 & 76.7 & \underline{85.8} \\
$\quad$SD ($\pm \sigma$) & 1.3 & 1.3 & 0.2 & 1.8 & 0.8 \\
+ LatentDR (Pool) & 86.3 & 82.6 & 97.1 & 79.2 & \textbf{86.3} \\
$\quad$SD ($\pm \sigma$) & 1.1 & 1.4 & 0.4 & 1.7 & 0.9 \\
\hline
\end{tabular}
% \end{sc}
\end{small}
\end{center}
\caption{Full results on PACS.}
\label{appendix:bd_pacs}
\end{table}

\paragraph{VLCS}
The breakdown performances are shown in Table \ref{appendix:bd_vlcs}, where $\sigma$ is the standard deviation (SD).

\begin{table}[h]
\begin{center}
\label{table-pacs}
\begin{small}
% \begin{small}
% \begin{sc}
\begin{tabular}{lcccccr}
\hline
 & C & L & S & V & Avg. \\
\hline
ERM & 96.2 & 63.9 & 72.2 & 74.4 & 76.7 \\
\hline
+ Mixup & 97.7 & 64.8 & 70.9 & 74.2 & 76.9 \\
% + I-Mixup$^{\dagger}$ &  &  &  &  & 84.6 \\
+ CutMix & 95.8 & 62.7 & 71.0 & 70.0 & 74.9 \\
+ Manifold-Mixup & 98.4 & 62.1 & 75.1 & 75.7 & 77.8 \\
+ MixStyle$^{\dagger}$ & 98.6 & 64.5 & 72.6 & 75.7 & 77.9 \\
+ BatchFormer & 97.0 & 64.5 & 70.9 & 74.5 & 76.7 \\
\hline
+ LatentDR (SA) & 97.8 & 64.5 & 73.9 & 78.4 & \textbf{78.7} \\
$\quad$SD ($\pm \sigma$) & 1.0 & 0.9 & 1.6 & 1.6 & 0.7 \\
+ LatentDR (Pool) & 98.0 & 66.2 & 69.4 & 78.4 & \underline{78.0} \\
$\quad$SD ($\pm \sigma$) & 0.6 & 1.2 & 1.0 & 1.7 & 0.5 \\
\hline
\end{tabular}
% \end{sc}
% \end{small}
\end{small}
\end{center}
\caption{Full results on VLCS.}
\label{appendix:bd_vlcs}
\end{table}

% \subsubsection{Performance on Office-Home}

\paragraph{Office-Home}
The breakdown performances are shown in Table \ref{appendix:bd_oh}, where $\sigma$ is the standard deviation (SD).

\begin{table}[h]
\begin{center}
\label{table-pacs}
\vskip 0.15in

\begin{small}
% \begin{sc}
\begin{tabular}{lcccccr}
\hline
 & A & C & P & R & Avg. \\
\hline
ERM & 58.8 & 52.0 & 73.3 & 75.1 & 64.8 \\
\hline
+ Mixup & 61.8 & 53.3 & 75.6 & 77.2 & 67.0 \\
% + I-Mixup$^{\dagger}$ &  &  &  &  & 84.6 \\
+ CutMix & 61.2 & 52.7 & 77.0 & 76.6 & 66.9 \\
+ Manifold-Mixup & 63.0 & 54.4 & 75.4 & 76.8 & 67.4 \\
+ MixStyle$^{\dagger}$ & 51.1 & 53.2 & 68.2 & 69.2 & 60.4 \\
+ BatchFormer & 60.0 & 51.9 & 74.5 & 76.7 & 65.8 \\
\hline
+ LatentDR (SA) & 65.3 & 54.9 & 77.3 & 78.5 & \textbf{69.0} \\
$\quad$SD ($\pm \sigma$) & 0.5 & 1.0 & 0.7 & 0.5 & 0.3 \\
+ LatentDR (Pool) & 63.6 & 56.1 & 75.6 & 78.2 & \underline{68.4} \\
$\quad$SD ($\pm \sigma$) & 0.5 & 0.4 & 0.8 & 0.6 & 0.3 \\
\hline
\end{tabular}
% \end{sc}
\end{small}
\end{center}
\caption{Full results on Office-Home.}
\label{appendix:bd_oh}
\end{table}

% \subsubsection{Performance on TerraIncognita}
\paragraph{TerraIncognita}
The breakdown performances are shown in Table \ref{appendix:bd_ti}, where $\sigma$ is the standard deviation (SD).

\begin{table}[h]
\begin{center}
\label{table-pacs}
\begin{small}
% \begin{sc}
\begin{tabular}{lcccccr}
\hline
 & L100 & L38 & L43 & L46 & Avg. \\
\hline
ERM & 54.2 & 41.0 & 55.4 & 37.5 & 47.0 \\
\hline
+ Mixup & 59.8 & 41.9 & 56.3 & 33.9 & 48.0 \\
% + I-Mixup$^{\dagger}$ &  &  &  &  & 84.6 \\
+ CutMix & 63.5 & 50.1 & 60.5 & 34.3 & \textbf{52.1} \\
+ Manifold-Mixup & 57.4 & 41.4 & 55.3 & 31.1 & 46.3 \\
+ MixStyle$^{\dagger}$ & 54.3 & 34.1 & 55.9 & 31.7 & 44.0 \\
+ BatchFormer & 54.0 & 43.4 & 56.0 & 41.0 & 48.6 \\
\hline
+ LatentDR (SA) & 49.6 & 47.1 & 58.2 & 44.2 & \underline{49.8} \\
$\quad$SD ($\pm \sigma$) & 2.6 & 2.8 & 1.1 & 2.5 & 1.5 \\
+ LatentDR (Pool) & 57.6 & 46.8 & 58.8 & 34.8 & 49.5 \\
$\quad$SD ($\pm \sigma$) & 3.6 & 3.0 & 1.0 & 2.2 & 1.9 \\
\hline
\end{tabular}
% \end{sc}
\end{small}
\end{center}
\caption{Full results on TerraIncognita.}
\label{appendix:bd_ti}
\end{table}

% \subsubsection{Performance on DomainNet}
\paragraph{DomainNet}
The breakdown performances are shown in Table \ref{appendix:bd_dnet}.

\begin{table*}[h]
\begin{center}
\label{table-pacs}
% \begin{small}
% \begin{sc}
\begin{tabular}{lcccccccr}
\hline
 & clipart & infograph & painting & quickdraw & real & sketch & Avg. \\
\hline
ERM & 60.4  & 19.2 & 48.1 & 12.4 & 60.4 & 50.9 & 41.9 \\
\hline
+ Mixup & 62.1 & 20.9 & 49.1 & 14.5 & 60.2 & 51.2 & 43.0 \\
% + I-Mixup$^{\dagger}$ &  &  &  &  & 84.6 \\
+ CutMix & 61.7 & 20.6 & 49.6 & 13.7 & 61.9 & 51.6 & 43.2 \\
+ Manifold-Mixup & 62.4 & 20.8 & 49.1 & 13.4 & 60.7 & 51.5 & 43.0 \\
+ MixStyle$^{\dagger}$ & 51.9 & 13.3 & 37.0 & 12.3 & 46.1 & 43.4 & 34.0 \\
+ BatchFormer & 62.6 & 19.4 & 48.5 & 13.2 & 61.8 & 51.6 & 42.8 \\
\hline
+ LatentDR (SA) & 63.6 & 22.3 & 51.5 & 14.6 & 64.7 & 54.0 & \textbf{45.1} \\
+ LatentDR (Pool) & 61.1 & 21.1 & 50.6 & 15.3 & 62.8 & 52.5 & \underline{43.9} \\
\hline
\end{tabular}
% \end{sc}
% \end{small}
\end{center}
\caption{Full results on DomainNet.}
\label{appendix:bd_dnet}
\end{table*}

\subsection{Additional ablations}

\paragraph{Sharing classifier weights.}
Our model trains a classifier $g$ on top of the encoder $f$, and uses the same classifier $g$ to regularize the model to learn degraded samples and restored samples. Following \cite{hou2022batchformer}, we tested whether training two different classifiers, one for the original loss, and the other for the corruption/restoration steps, would benefit the training. With a separate classifier, our model gets an average of $83.7\%$ accuracy on PACS, an over 2\% decrease. It seems it is critical to share classifier weights for our method. We hypothesize that this is because
%the fundamental difference between \ours~and BatchFormer: 
BatchFormer relies on the gradient flow across samples to guide learning, while \ours~uses latent augmentations to regularize training.

\paragraph{Increasing the batch size.}

\begin{figure}
\begin{center}
 \includegraphics[width=0.38\textwidth]{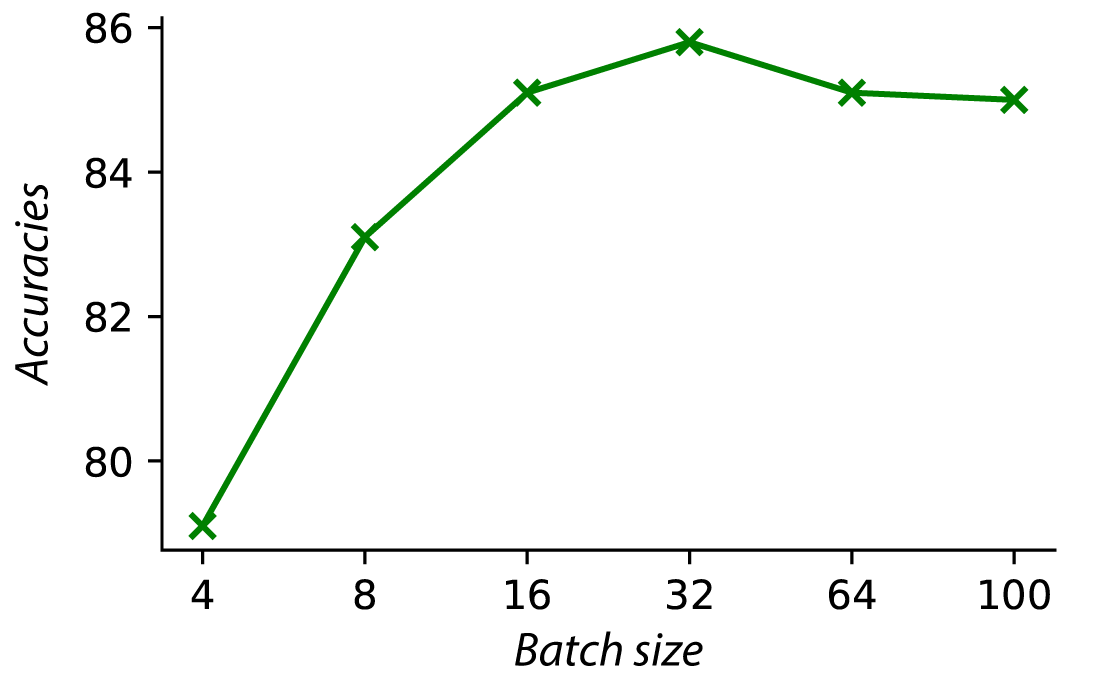}
 \end{center}
   \caption{\textit{Ablation on batch size.} We evaluate the performance of \ours~(SA) with different batch size on the PACS dataset.}
 \label{fig:app_bs}
\end{figure}

To understand if our method would benefit from learning with larger batch size, we evaluate the performance of \ours~on the PACS dataset with $B = [4, 8, 16, 32, 64, 100]$, where each training domain provides $B$ samples for training (thus, the batch size is $3B$ for the PACS dataset). As shown in Figure~\ref{fig:app_bs}, training \ours~requires a sufficiently large batch size to capture rich and meaningful sample-to-sample relationship across different classes and domains.
% results are visualized in .
However, further increasing the batch size does not further increase the performance of our method. We hypothesize that it might be challenging for the small-scale transformer to capture information across too many tokens inside the same batch.

% Reason: As PACS consists of 7 classes, when we are using BS 4, the mini-batch will not contain even 1 sample from all the classes for each of the domains. Due to this reason, we observe a significant jump in accuracy from BS 4 to BS 8 in both self-attention and one-2-many. The extra number of samples per class seems to have a positive effect on self-attention but does not change the performance of one-2-many by much.

% \include{texts/additional_backbone}

% \paragraph{Combining BT with better optimizer:}
% In addition,
% \colortext{The results are not as impressive when combining BT with another optimizer. Terra results are much better, the other datasets are more or less similar}
% to testing the performance of BT on its own, we combine BT with other existing methods outside of ERM. When designing this, we specifically tend to select methods that do not have too much latent space augmentation already, and thus we selected SWAD.

\subsection{Additional visualizations}
\label{appendix:latents}

\begin{figure}[t]
\begin{center}
\includegraphics[width=0.49\textwidth]{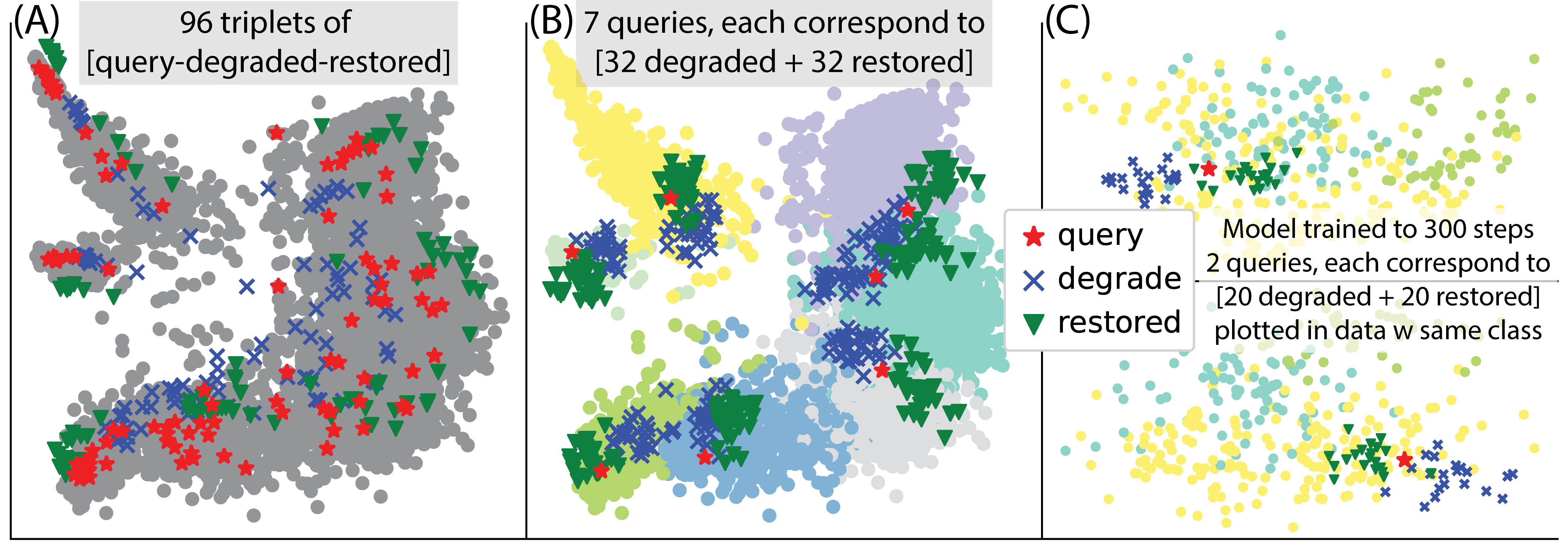}
\end{center}
   \caption{{Latent queries} ({\color{red}{$\star$}}), the {degraded} ({\color{blue}{$\times$}}), and the {restored} ({\color{darkergreen}{$\blacktriangledown$}}). The left two figures show all training data as background.
   }
\label{fig:all}
\end{figure}

\paragraph{Additional details and copies of Figure 1}~
We provide additional visualizations as in Figure~\ref{fig:all} as additional evidence to Figure~\ref{fig:fig1}.
Figure~\ref{fig:all}(A) shows a random batch (96 samples) of queries ({\color{red}{$\star$}}), their degraded pairs ({\color{blue}{$\times$}}), and their restored pairs ({\color{darkergreen}{$\blacktriangledown$}}) in one forward pass. We note that the restored latents typically are
% \zkn{What is this both referring to?} 
{\textbf{far away from the original latents}}.
In Figure~\ref{fig:all}(B), we show how we generated Figure~\ref{fig:fig1}(B): For each random query ({\color{red}{$\star$}}) selected, we insert it into 32 random batches from the training data, and plot both their degraded and restored latents on top of the original representations.
In Figure~\ref{fig:all}(C), we generate Figure~\ref{fig:fig1}(A) using a model that is trained on PACS for 300 steps, with training data from the same class as the background. 

\paragraph{Visualizations on the testing domain}

\begin{figure}
\begin{center}
 \includegraphics[width=0.46\textwidth]{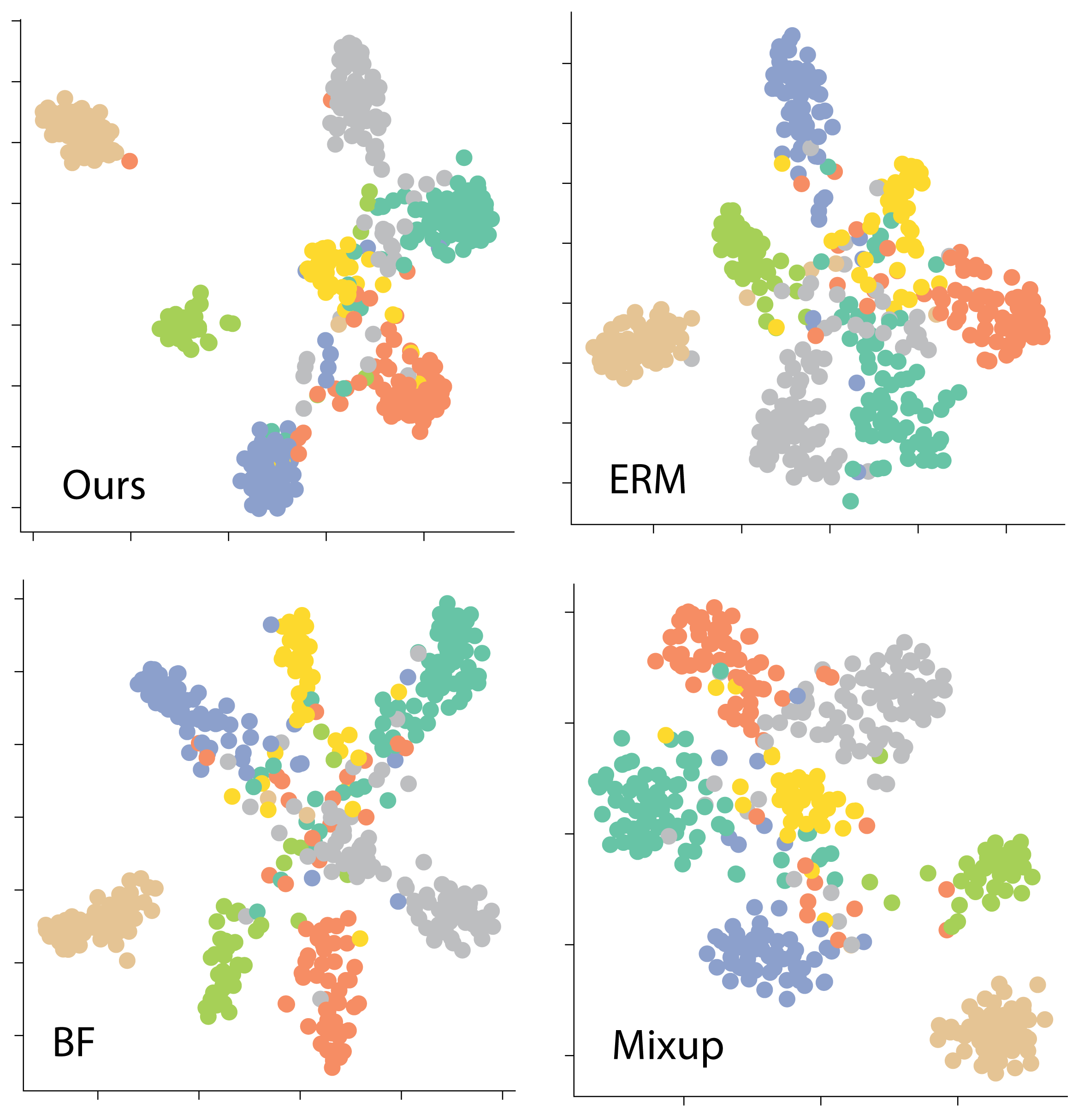}
 \end{center}
   \caption{\textit{Latent space direct visualization.} We visualize the latent spaces for ERM, Mixup \cite{zhang2017mixup}, BatchFormer \cite{hou2022batchformer} (BF), and \ours~(Ours-SA) using a T-SNE on the testing domain. Our model provides the best alignment (closeness of latents from the same class) and the best uniformity (most information from the training data is preserved).}
 \label{fig:app_latents}
 \vspace{-2mm}
\end{figure}

We provide the direct visualization of latent space as in Figure~\ref{fig:app_latents}, where we use a T-SNE to visualize the model's latent on the testing domain (domain A) for models that are trained on the PACS dataset. Different color represents different classes for the classification task on PACS. As shown in Figure~\ref{fig:app_latents}, \ours~provides the most clustered latents for each class in comparison to other methods, while for each class, \ours~provides clusters with the most circular shape. These properties are further demonstrated through the measurement of alignment and uniformity, where our method achieves the best `closeness' score for classification, and the best latent `diversity' score.

%\vspace{-2mm}

\end{document}